\documentclass[10pt,twocolumn,letterpaper]{article}

\usepackage[pagenumbers]{cvpr} 

\usepackage[dvipsnames]{xcolor}

\definecolor{cvprblue}{rgb}{0.21,0.49,0.74}
\usepackage[pagebackref,breaklinks,colorlinks,citecolor=cvprblue]{hyperref}

\def\paperID{***} 
\def\confName{CVPR}
\def\confYear{2024}

\definecolor{darkgreen}{rgb}{0.55, 0.74, 0.42}

\usepackage{pifont}
\newcommand{\cmark}{\ding{51}}%
\newcommand{\xmark}{\ding{55}}%

\title{Stream Query Denoising for Vectorized HD Map Construction}

\author{Shuo Wang\textsuperscript{1$\dagger$}  
\quad Fan Jia\textsuperscript{2} 
\quad Yingfei Liu\textsuperscript{2} 
\quad Yucheng Zhao\textsuperscript{2}  
\quad Zehui Chen\textsuperscript{1}
\\
Tiancai Wang\textsuperscript{2}
\quad Chi Zhang\textsuperscript{2}
\quad Xiangyu Zhang\textsuperscript{2}
\quad Feng Zhao\textsuperscript{1}  
\\
\textsuperscript{1}University of Science and Technology of China
\quad
\textsuperscript{2}MEGVII Technology  
}

\usepackage{multirow}
\usepackage{graphicx}
\usepackage{color}
\usepackage{colortbl}
\begin{document}
\maketitle
\setlength{\skip\footins}{0.4cm}
\renewcommand{\thefootnote}{\fnsymbol{footnote}} 
\footnotetext{$\dagger$ Work done was during the internship at MEGVII Technology.} 
\begin{abstract}
To enhance perception performance in complex and extensive scenarios within the realm of autonomous driving, there has been a noteworthy focus on temporal modeling, with a particular emphasis on streaming methods. The prevailing trend in streaming models involves the utilization of stream queries for the propagation of temporal information. Despite the prevalence of this approach, the direct application of the streaming paradigm to the construction of vectorized high-definition maps (HD-maps) fails to fully harness the inherent potential of temporal information.
This paper introduces the Stream Query Denoising (SQD) strategy as a novel approach for temporal modeling in high-definition map (HD-map) construction. SQD is designed to facilitate the learning of temporal consistency among map elements within the streaming model. The methodology involves denoising the queries that have been perturbed by the addition of noise to the ground-truth information from the preceding frame. This denoising process aims to reconstruct the ground-truth information for the current frame, thereby simulating the prediction process inherent in stream queries.
The SQD strategy can be applied to those streaming methods (e.g., StreamMapNet) to enhance the temporal modeling. The  proposed SQD-MapNet is the StreamMapNet equipped with SQD.
Extensive experiments on nuScenes and Argoverse2 show that our method is remarkably superior to other existing methods across all settings of close range and long range. The code will be available soon. 
\end{abstract}    
\section{Introduction}
\label{sec:intro}
The High-Definition Map (HD-map) serves the crucial purpose of furnishing centimeter-level location information for map elements and plays a pivotal role in various applications within autonomous driving, including localization \cite{wang2023multi, yurtsever2020survey, wang2023towards, liu2022petr, wu20231st} and navigation \cite{antonello2017fast, bekir2007introduction, hasan2009review}. Traditionally, the construction of HD-maps is conducted offline through SLAM-based methods \cite{zhang2014loam, shan2018lego}, which is both time-consuming and labor-intensive.
Recent research endeavors have shifted towards the construction of local maps within a predetermined range using onboard sensors. Although many existing works frame map construction as a semantic segmentation task \cite{zhang2022beverse, peng2023bevsegformer, qin2023unifusion, li2022hdmapnet, liu2023petrv2}, rasterized representations in such approaches exhibit redundant information, lack structural relationships between map elements, and often require extensive post-processing efforts \cite{li2022hdmapnet}. In response to these limitations, MapTR \cite{liao2022maptr} adopts an end-to-end approach to construct vectorized maps, akin to the DETR paradigm \cite{carion2020end, zhu2020deformable, liu2022dab, chang2023detrdistill}.

\begin{figure}[!t]
\setlength{\abovecaptionskip}{0cm}  
\setlength{\belowcaptionskip}{-0.3cm} 
  \centering
   \includegraphics[width=1.0\linewidth]{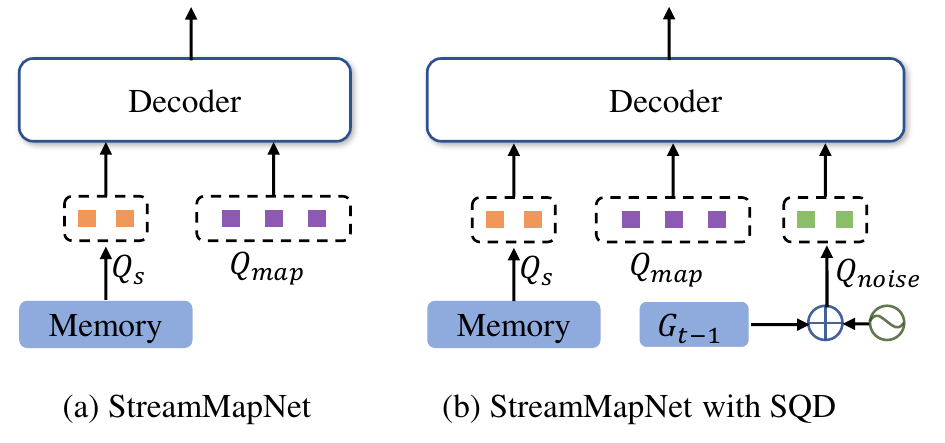}
   \vspace{-0.9em}
   \caption{
  (a). The decoder process of StreamMapNet~\cite{yuan2023streammapnet}. (b). StreamMapNet with the proposed stream query denoising (SQD).
   \textcolor{violet}{Purple blocks} are global map queries learned by the network, \textcolor{orange}{orange blocks} represent the stream queries from memory cache, and \textcolor{darkgreen}{green blocks} are the noised queries generated by adding noise to the ground truth $G_{t-1}$ of previous frame $t-1$. 
}
   \label{fig:teaser}
\end{figure}

Nevertheless, the aforementioned methods overlook the incorporation of temporal information. The efficacy of propagating sparse queries (hidden states) from the previous frame to the current frame has been demonstrated in temporal multi-view 3D object detection \cite{wang2023exploring, lin2023sparse4d}. While recent approaches grounded in vectorized representations share a similar paradigm with object detection, the direct application of the aforementioned temporal methods is not warranted due to inherent modeling variability between curves and bounding boxes.
In the context of object detection, determining the speed of the ego and surrounding objects enables the prediction of their positions at the next timestamp. This stands in contrast to scenarios involving lines, where changes over time result in new parts appearing and old parts detaching from the line, presenting a distinctive challenge not encountered in object detection.

\begin{figure}[!t]
\setlength{\abovecaptionskip}{0cm}  
\setlength{\belowcaptionskip}{-0.3cm} 
  \centering
   \includegraphics[width=1.0\linewidth]{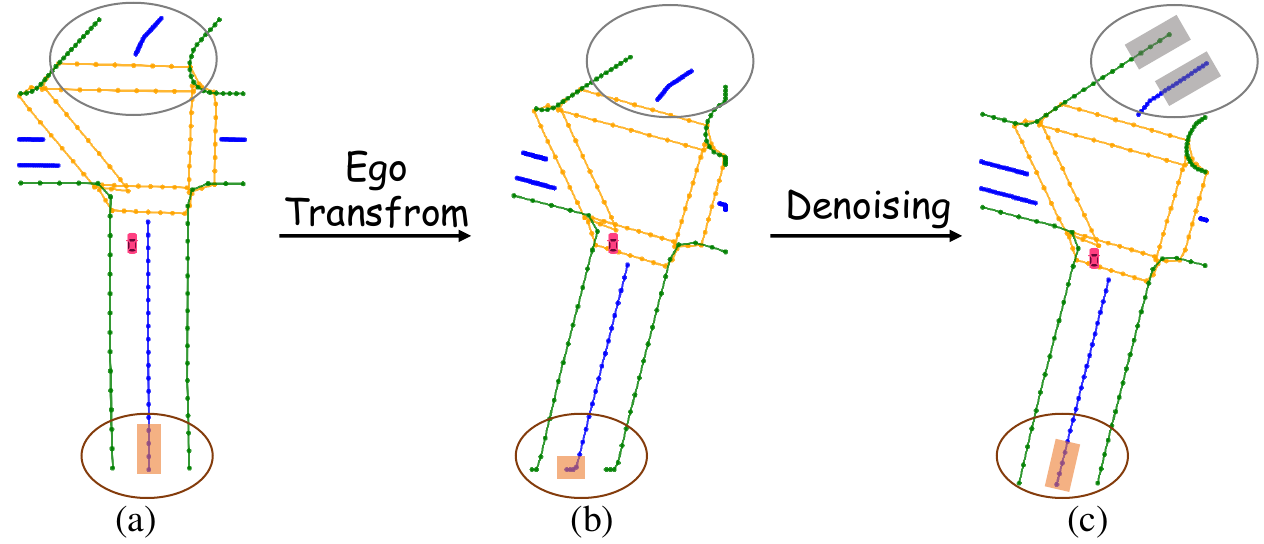}
   \vspace{-0.9em}
   \caption{(a) shows the ground truth of previous frame. (b) is the transformation of (a) to current frame according to ego motion. (c) is the ground truth of current frame.}
   \label{fig:intro}
\end{figure}

Suppose the model's predictions at the preceding moment precisely match the ground truth, and this valuable information is propagated to the current moment for improved initialization. Owing to ego-motion, the predictions must undergo transformation based on the matrix representing the transition between two frames. As depicted in \cref{fig:intro}, (a) and (b) illustrate the curves before and after transformation, respectively. The gray segment signifies the newly added part of the curve, necessitating all points along the entire line to acquire distinct offsets to accommodate the curve's growth.
Simultaneously, due to alterations in the perceptual range between the two frames, numerous points from the prior frame are truncated to the boundary of the current range, as depicted in the orange portion of \cref{fig:intro}. Points in this segment must assimilate different biases despite originating from nearly the same boundary. Clearly, explicitly teaching a network to grasp such intricate and diverse changes poses a challenge, and the temporal learning process for a network runs counter to conventional training. 

To address the aforementioned challenges, we propose a novel approach called Stream Query Denoising (SQD) for the streaming model, exemplified by StreamMapNet~\cite{yuan2023streammapnet}, in the context of HD-map construction. Illustrated in \cref{fig:teaser} (a), StreamMapNet utilizes stream queries as hidden states for propagating temporal information. Our SQD strategy is designed to facilitate the learning of temporal consistency among map elements. The denoising process, involving the addition of noise to $G_{t-1}$ to generate noise queries and subsequent reconstruction to simulate the prediction process of stream queries, is depicted in \cref{fig:teaser} (b).
The SQD strategy consists of two main components. First, we introduce normal query denoising, considering three distinct noise strategies for curves: line shifting, angular rotation, and scale transformation. The acquisition of noise queries tailored for map elements is further explored. Second, for stream query denoising, we incorporate temporal adaptive matching to establish an explicit one-to-one correspondence between historical ground truths and current ones. To address warped errors inherent in temporal and current elements, a dynamic query noising mechanism is devised.

In summary, the primary contributions of this paper are outlined as follows:
\begin{itemize}
\item[$\bullet$] The exploration of normal query denoising for HD-map construction, marking the first instance of such investigation. This encompasses three distinct noise strategies tailored for curves, accompanied by a detailed exploration of the methodology for acquiring noise queries.
\item[$\bullet$] We propose the stragegy of stream query denoising to assist the streaming model learn the temporal consistency of map elements.
\item[$\bullet$] The development of SQD-MapNet, integrating our SQD strategy into StreamMapNet. Results showcase its notable superiority over state-of-the-art methods on existing benchmarks in both original and novel settings, underscoring the effectiveness of our proposed approach.
\end{itemize}


\section{Related Works}
\label{sec:relate}

\subsection{Online Vectorized HD-Map Construction}
There has been a surge of interest in leveraging onboard sensors for the construction of vectorized local HD-maps. HDMapNet~\cite{li2022hdmapnet} employs a semantic map prediction approach, followed by the aggregation of pixel-wise segmentation results through post-processing. In an effort to mitigate redundant information and alleviate the need for time-consuming post-processing, VectorMapNet~\cite{liu2023vectormapnet} introduces a refinement step for map elements using an auto-regressive transformer. MapTR~\cite{liao2022maptr} adopts hierarchical queries and a fixed number of points to represent the map, while BeMapNet~\cite{qiao2023end} utilizes piecewise Bezier curves to model map elements. Additionally, PivotNet~\cite{ding2023pivotnet} presents a map construction method based on pivot-based representations.

\subsection{Temporal Camera-based Perception}
The significance of temporal information is paramount, especially in intricate scenarios involving long distances, occlusion, and the like. Notably, the utilization of temporal information has been a focal point in the domain of camera-based 3D object detection. Approaches such as BEVDet4D~\cite{huang2022bevdet4d} and BEVFormer v2~\cite{yang2023bevformer} adopt the strategy of stacking features from multiple historical frames and processing them in a single forward pass. However, this method incurs substantial computational costs and imposes limitations on the number of historical frames that can be effectively utilized. In contrast, VideoBEV~\cite{han2023exploring} incorporates a recurrent long-term fusion module to sequentially fuse BEV features in a video stream. StreamPETR~\cite{wang2023exploring} and Sparse4D v2~\cite{lin2023sparse4d} introduce the streaming queries strategy to propagate temporal information. Notably, StreamMapNet~\cite{yuan2023streammapnet} extends this core idea to the construction of HD-Map by applying streaming queries and streaming BEV features.

\begin{figure*}[h]
  \centering
   \includegraphics[width=1.0\linewidth]{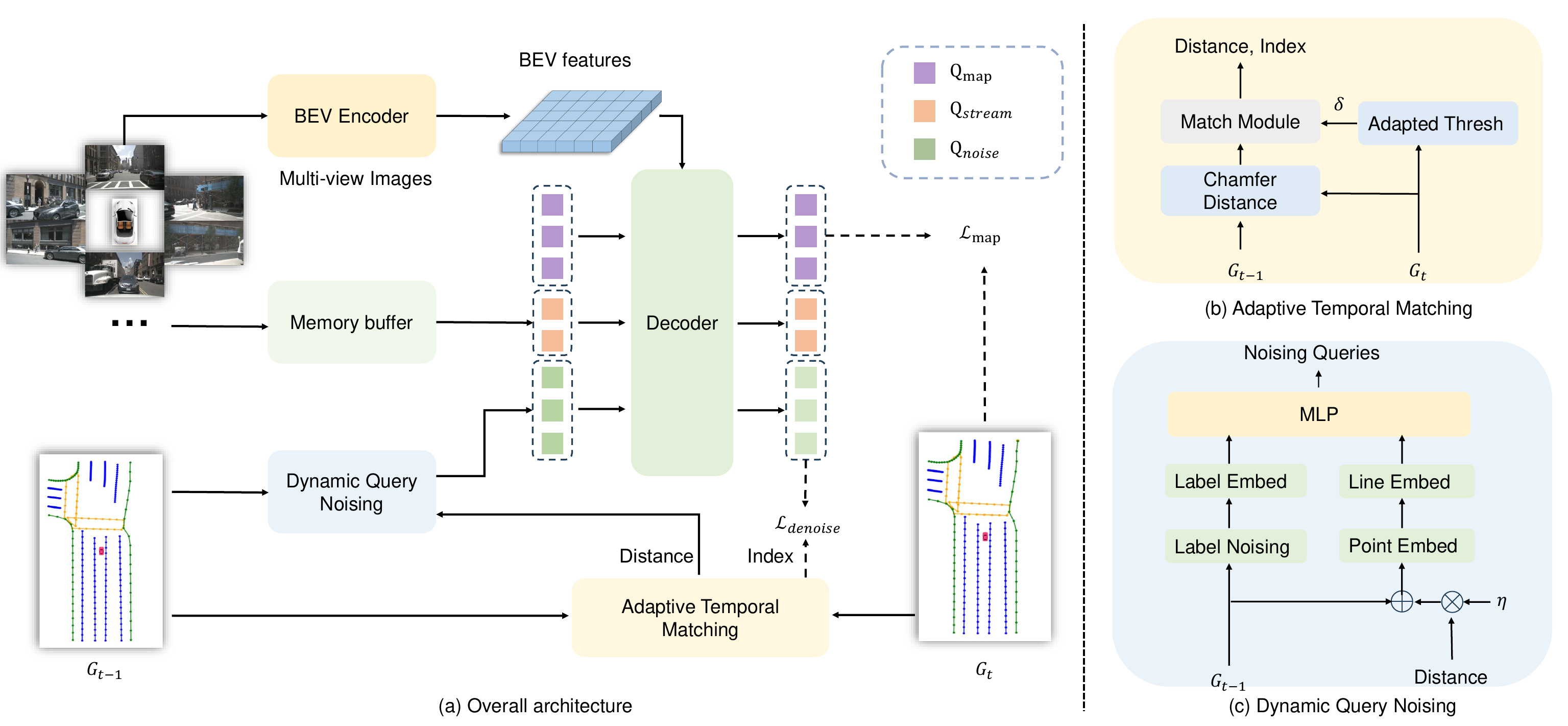}
   \caption{(a) shows the overall framework of SQD-MapNet. (b) and (c) are the specific implementations of adaptive temporal matching and dynamic query noising, respectively. $G_{t-1}$ is the ground-truth of the last frame $t-1$.
}
   \label{fig:network}
   \vspace{-1.0em}
\end{figure*}

\subsection{Query Denoising}
DN-DETR~\cite{li2022dn} pioneers the utilization of query denoising to address the instability inherent in bipartite graph matching. DINO~\cite{zhang2022dino} extends this concept by introducing a definition of negative samples based on DN-DETR. Furthering this approach, MaskDINO~\cite{li2023mask} performs object detection and segmentation tasks concurrently. DN-MOT~\cite{fu2023denoising} tailors a denoising strategy to mitigate the impact of occlusion in multiple object tracking. In the context of this paper, we introduce the concept of stream query denoising for temporal HD-map construction. To the best of our knowledge, this is the first instance where the efficacy of query denoising in the temporal modeling of HD-map construction has been systematically examined.

\section{Methodology}
\label{sec:method}
We first review the pipeline of StreamMapNet, followed by an exploration of normal query denoising. Subsequently, we introduce our approach to stream query denoising, elucidating two specific modules: adaptive temporal matching and dynamic query noising. The comprehensive architecture of our approach is illustrated in \cref{fig:network}.

\subsection{A Review of StreamMapNet}
The overall pipeline of StreamMapNet consists of three primary components as follows:

\noindent\textbf{BEV Feature Encoder.} Given multi-view images from the onboard cameras, a shared backbone is employed to extract image features. Then these features are aggregated and processed by a Feature Pyramid Network~\cite{lin2017feature} (FPN). Finally, various PV2BEV transformation methods~\cite{li2022bevformer, philion2020lift} can be applied to obtain the BEV feature. Without loss of generality, we adapt the way of BEVFormer~\cite{li2022bevformer} in this work.

\noindent\textbf{Polyline Decoder.} Similar to most works ~\cite{liao2022maptr, liu2023vectormapnet,ding2023pivotnet}, StreamMapNet views the vectorized map construction as a set prediction task and utilizes an end-to-end decoder ~\cite{carion2020end} to extract lines from the BEV features, \ie, the module takes the BEV feature and a set of learnable instance queries $\{Q_m\}_{m=1}^M$, and outputs a set of point descriptors$\{D_m\}_{m=1}^M$, where $m$ is the max number of instances. Specifically, queries first perceive the global scene through global attention. Then each query generates $n$ reference points, where $n$ denotes the number of points forming a curve. Next, multi-point deformable attention of queries is performed on the dense BEV feature. Finally, we can obtain the final position of point descriptors $D_m=\{p_1^m, p_2^m, \cdots, p_n^m\}$.

\noindent\textbf{Temporal Query Propagation.} Due to the static nature of map elements, there is a high probability that the instance at the current moment will continue to appear at the next moment, which means that the queries at present can provide a better reference position for the next moment than global initialization. StreamMapNet adopts the query propagation and BEV fusion to perform temporal fusion. Concretely, for query propagation, queries with the highest $k$ scores as $Q_{t-1}$ at the current frame are selected and the corresponding predicted reference points $P_{t-1}$ can be obtained. Considering the movement of the ego, it utilizes the transformation matrix $T$ between the coordinate systems of two frames before propagation as
\begin{gather}
  Q_t^{\prime} = \phi _t({\rm Concat}(Q_{t-1}, {\rm Flatten}(T))) + Q_{t-1} \\
  P_t^{\prime} = T \cdot  P_{t-1}
\end{gather}
To this end,  $Q_t^{\prime}$ and $P_t^{\prime}$ can be propagated to the next frame. Moreover, StreamMapNet employs a Gated Recurrent Unit~\cite{chung2014empirical} (GRU)  to fuse these temporal BEV features.

\begin{figure}[t]
\setlength{\abovecaptionskip}{0cm}  
\setlength{\belowcaptionskip}{-0.3cm} 
  \centering
   \includegraphics[width=1.0\linewidth]{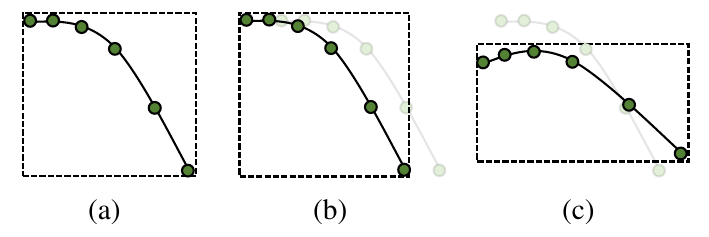}
   \vspace{-15pt}
   \caption{(a) is the original curve, which consists of a number of points and is surrounded by the minimum bounding rectangle. (b) and (c) shows box shifting and box scaling, respectively. The light-colored curves indicate the curves before noise addition.}
   \label{fig:noise}
\end{figure}

\subsection{Normal Query Denoising}
The DETR series~\cite{carion2020end, zhu2020deformable, liu2022dab} faces challenges stemming from the instability of bipartite graph matching, leading to slow convergence and suboptimal results. Addressing these issues, DN-DETR~\cite{li2022dn} introduces the strategy of query denoising, effectively mitigating the aforementioned problems.
However, as bounding boxes and curves exhibit inherent differences, the direct application of DN-DETR to the context of HD-map construction is not feasible. Consequently, we propose a novel approach termed normal query denoising. In this regard, we initially explore noise strategies tailored for curves. Subsequently, we detail the methodology for acquiring noise queries specific to HD-map construction.

\subsubsection{Noise Strategies for Curves}
\label{3.3.2}
A pivotal aspect of the denoising process hinges on the design of the noise, and to date, no prior work has delved into noise strategies tailored for curves. In this section, we examine three distinct noise strategies for curves: \emph{line shifting}, \emph{angular rotation}, and \emph{scale transformation}. Here, line shifting and angular rotation refer to the global translation and rotation of the curve, respectively, while scale transformation pertains to the scaling of the curve's length.

Inspired by the principles of DN-DETR \cite{li2022dn}, we introduce a novel approach that unifies the characterization of curves and bounding boxes for the first time. We propose a simple yet effective method for incorporating the three aforementioned noise types. Specifically, we encapsulate the curve by considering its minimum bounding rectangle as an instance representation, as depicted in \cref{fig:noise} (a). Assuming $x$ and $y$ represent the center coordinates of the bounding box, and $w$ and $h$ denote the corresponding width and height, the noise strategies for curves can be implemented in two distinct ways.

1) Box shifting: we introduce a random shift $(\Delta x, \Delta y)$ to the box, subject to the constraint $|\Delta x|<\frac{\lambda_1w}{2}$ and $|\Delta y|<\frac{\lambda_2h}{2}$, where $\lambda_1,$ $\lambda_2$ are the maximum scale of noise. The position of the points in the curve with respect to the bounding box remains unchanged. This achieves the effect of \emph{line shifting}, as illustrated in \cref{fig:noise} (b).
2) Box scaling: We randomly sample the height and width within the intervals $[(1-\lambda_3)h, (1+\lambda_3)h]$ and $[(1-\lambda_4)w, (1+\lambda_4)w]$, respectively, where $\lambda_3$ and $\lambda_4$ denote the maximum scale of noise. Despite alterations in the angle and length of the curve, the relative positions of the points with respect to the box remain unchanged. Consequently, this approach achieves \emph{angular rotation} and \emph{scale transformation} simultaneously, as depicted in \cref{fig:noise} (c).

\subsubsection{Acquisition of Noising Query}
\label{acquire query}
Given a noisy curve, its category and point sets are denoted as $cls$ and $\{p_1^m, p_2^m, \cdots, p_n^m\}$, where $m$ is the index of the curve and $n$ is the number of points forming an instance. A learnable embedding is set for each category and then we can acquire the content embedding $C_q \in \mathbb{R}^{\frac{D}{2}}$, where $D$ is the dimension of decoder embedding. For a point $p_i^m=\{x_i^m, y_i^m\}$, its position embedding can be generated by
\begin{equation}
      {P}_i^m = {\rm MLP}^{(pt)}({\rm Concat}({\rm PE}(x_i^m), {\rm PE}(y_i^m))).
\end{equation}
In our implementations, the positional encoding function PE maps a float to a vector with $\frac{D}{4}$ dimensions as: PE: $\mathbb{R} \rightarrow \mathbb{R}^{\frac{D}{4}}$, and the function MLP$^{(pt)}$ projects a $\frac{D}{2}$ dimensional vector into $\frac{D}{2}$ dimensions: MLP$^{(pt)}$: $\mathbb{R}^{\frac{D}{2}} \rightarrow \mathbb{R}^{\frac{D}{2}}$.
Then the position embedding of the instance can be obtained as 
\begin{equation}
      {Pos}_q = {\rm MLP}^{(pos)}({\rm Concat}({P}_1^m, {P}_2^m, \cdots, {P}_n^m)),
\end{equation}
where MLP$^{(pos)}$: $\mathbb{R}^{\frac{nD}{2}} \rightarrow \mathbb{R}^{\frac{D}{2}}$ fuses the information from all the points into the position information of the curve.

Since we use deformable attention to interact the queries with the reference points for information, we acquire the denoising query $Q_{denoise}$ by fusing the content information with the position information via MLP$^{(fuse)}$: $\mathbb{R}^{D} \rightarrow \mathbb{R}^{D}$ as 
\begin{equation}
      Q_{denoise} = {\rm MLP}^{(fuse)}({\rm Concat}(C_q, Pos_q)),
\end{equation}

\subsection{Stream Query Denoising}
Query denoising (DN) \cite{li2022dn} has demonstrated its efficacy in improving the network's ability to model queries. However, as illustrated in \cref{fig:intro}, the intricate nature of predictions generated by streaming queries introduces challenges when adding noise to the current ground truth. This complexity arises from the difficulty in simulating both the increasing and decreasing parts of previously predicted curves.
To address this, we introduce Stream Query Denoising (SQD), a novel approach that involves adding noise to the ground truth of the previous frame to emulate the predictions made by streaming queries. Moreover, considering that a portion of curves in the previous ground truth remains unchanged due to ego-motion, adding noise to the ground truth of the previous frame is tantamount to DN.
Therefore, SQD not only facilitates the network in learning to model temporal queries and current queries simultaneously but also aligns with the principles of DN, making it a versatile strategy for enhancing the network's temporal understanding.
Next, we will introduce two essential components, namely adaptive temporal matching and dynamic query noising, to effectively complement and enhance the Stream Query Denoising (SQD) strategy.

\subsubsection{Adaptive Temporal Matching}
\label{3.3.1}
In the process of stream query denoising, we have no access to an explicit one-to-one correspondence between temporal ground truth and current ones. In order to circumvent the instability associated with bipartite graph matching, we propose Adaptive Temporal  Matching (ATM). 

Let $\{y_{1}^{t-1}, y_{2}^{t-1}, \cdots, y_{m}^{t-1}\}$ and $\{y_{1}^{t}, y_{2}^{t}, \cdots, y_{n}^{t}\}$ represent the ground truth in the previous frame and the current frame, respectively. Here, $y$ is composed of a fixed number of points, and $m$ and $n$ denote the number of points in the respective ground truth. Due to the movement of the ego, the transformation matrix $T$ between the coordinate systems of the two frames is employed to convert $\{y_{1}^{t-1}, y_{2}^{t-1}, \cdots, y_{m}^{t-1}\}$ to $\{\hat{y}_{1}^{t-1}, \hat{y}_{2}^{t-1}, \cdots, \hat{y}_{m}^{t-1}\}$. For each curve in the current frame $y_{i}^{t}$, we compute the Chamfer distance (CD) between it and each instance in the previous frame. The minimum distance and its corresponding location are preserved::
\begin{gather}
  {\rm CD}_{Dir}(S_1, S_2) = \frac{1}{S_1} \sum_{p \in S_1} \mathop{{\rm min}}\limits_{q \in S_2} ||p-q ||_2 , \\
  {\rm CD}(S_1, S_2) = {\rm CD}_{Dir}(S_1, S_2) + {\rm CD}_{Dir}(S_2, S_1), \\
  D, idx = \mathop{{\rm min}}\limits_{j \in [1, m]}({\rm CD}(y_{i}^{t}, \hat{y}_{j}^{t-1})),
  \label{eq:8}
\end{gather}
where CD$_{Dir}$ is the directional Chamfer distance and CD is the bi-directional Chamfer distance; $S_1$ and $S_2$ are the two sets of points on the curves; $idx$ denotes the index closest to the curve itself and $D$ is the minimum distance value.

Due to the inherent differences in the properties of each curve, a unique tolerance threshold is set for each instance. Initially, we extract the maximum bounding rectangle from the curve and consider the width $w$ and height $h$ of the rectangle as the reference scale for this curve. Subsequently, the threshold is calculated as:
\begin{equation}
  \delta = \alpha \frac{w+h}{2},
  \label{eq:3}
\end{equation}
where $\alpha$ is the degree of the tolerance.

Only when the minimum distance in~\cref{eq:8} is less than the threshold of the corresponding instance in the current frame, we will assign the temporal instance to the current curve. Consequently, through Adaptive Temporal Matching (ATM), we establish the correspondence between the temporal ground truth and the current one.

\subsubsection{Dynamic Query Noising}
Upon transforming the ground truth of the previous frame into the current frame based on ego-motion, a natural bias emerges between it and the ground truth in the current frame. Perturbing all ground truths equally, without considering the instance's inherent noise, is suboptimal. To address this, we introduce Dynamic Query Noising.

After obtaining the matching relation in \cref{3.3.1}, the corresponding minimum Chamfer distance $D$ can also be derived from \cref{eq:8}. Subsequently, the decay rate of noise for each instance can be calculated as follows:
\begin{equation}
  R_{dacay} = 1 - \frac{D}{\gamma \cdot \frac{\delta}{\alpha}},
\end{equation}
where $\gamma$ is the predefined decay scale and $\delta$ is acquired by ~\cref{eq:3}.

To illustrate, considering the box-level scale in~\cref{3.3.2} as an example, if we assume that the randomly added noise is  $\eta =\{\Delta x, \Delta y, \Delta w, \Delta h\}$, then the final instance is represented as
\begin{equation}
  B_{ins} = \{x, y, w, h\} + \{\Delta x, \Delta y, \Delta w, \Delta h\} \cdot R_{decay},
\end{equation}
where $\{x, y, w, h\}$ is the original instance.

After getting the noisy categories and point sets, we derive the noising queries according to \cref{acquire query}.

\subsubsection{Objective Function}
Our model adopts an end-to-end training approach. During training, we employ the same map loss function as StreamMapNet\cite{yuan2023streammapnet}, including the classification loss $\mathcal{L}_{Focal}$, the line loss $\mathcal{L}_{line}$ and the translation loss $\mathcal{L}_{trans}$. 
\begin{gather}
  \mathcal{L}_{map} = \lambda_1 \mathcal{L}_{Focal} + \lambda_2 \mathcal{L}_{line} + \lambda_3 \mathcal{L}_{trans} , 
\end{gather}
where $\lambda_1, \lambda_2 $ and $\lambda_3$ are hyperparameters.

Additionally, for the prediction results of denoising queries, we use the same type of classification loss and line loss to construct $\mathcal{L}_{denoise}$. 
\begin{gather}
  \mathcal{L}_{denoise} = \lambda_4 \mathcal{L}^{DN}_{Focal} + \lambda_5 \mathcal{L}^{DN}_{line}, 
\end{gather}
where $\mathcal{L}^{DN}_{Focal}$ and $\mathcal{L}^{DN}_{line}$ are classification loss and the line loss of the denoising predictions and $\lambda_4, \lambda_5$ are hyperparameters.

Finally, the overall loss is defined as: 
\begin{gather}
  \mathcal{L}_{train} = \mathcal{L}_{map}  + \mathcal{L}_{denoise}
\end{gather}
\section{Experiments}
\label{sec:exper}

\begin{table*}
\setlength{\abovecaptionskip}{0.2cm}  
\setlength{\belowcaptionskip}{-0.2cm} 
  \centering
  \begin{tabular}{c|cccc|cccc}
    \toprule
     Range& Method & Backbone & Image Size & Epoch & AP$_{ped}$ & AP$_{div}$ & AP$_{bound}$ & mAP \\
    \midrule
    \multirow{8}*{60 $\times$ 30 \emph{$m$}} & VectorMapNet~\cite{liu2023vectormapnet} &R50  &256 $\times$ 480  &110  &36.1  &47.3  &39.3 &40.9 \\
    ~ & MapTR~\cite{liao2022maptr} &R50  &480 $\times$ 800  &24  &46.3  &51.5  &53.1 &50.3 \\
    ~ & BeMapNet~\cite{qiao2023end} &R50  &512 $\times$ 896  &30  &57.7  &62.3  &59.4 &59.8 \\
    ~ & PivotNet~\cite{ding2023pivotnet} &R50  &-  &24  &56.2  &56.5  &60.1 &57.6 \\
    ~ & StreamMapNet~\cite{yuan2023streammapnet} &R50  &480 $\times$ 800  &24  &60.4  &61.9  &58.9 &60.4 \\
    ~ & SQD-MapNet (Ours) &R50  &480 $\times$ 800  &24  &\textbf{63.0}  &\textbf{62.5}  &\textbf{63.3} &\textbf{63.9} \\
    \cline{2-9}
    ~ &  StreamMapNet$^\dag$~\cite{yuan2023streammapnet} &R50  &480 $\times$ 800  &24  &-  &-  &- &62.9 \\
    ~ & SQD-MapNet$^\dag$ (Ours) &R50  &480 $\times$ 800  &24  & \textbf{63.6}  &\textbf{66.6}  &\textbf{64.8} &\textbf{65.0} \\
    \rowcolor[gray]{.9} ~ &  SQD-MapNet (Ours) &V2-99  &900 $\times$ 1600  &110  &\textbf{74.2}  &\textbf{72.3}  &\textbf{75.6} &\textbf{74.0} \\
    \midrule
    \multirow{3}*{100 $\times$ 50 \emph{$m$}} &  MapTR~\cite{liao2022maptr} &R50  &480 $\times$ 800  &24  &45.5  &47.1  &43.9 &45.5 \\
    ~ & StreamMapNet~\cite{yuan2023streammapnet} &R50  &480 $\times$ 800 &24 &62.9 &63.1  &55.8  &60.6 \\
    ~ & SQD-MapNet (Ours) &R50  &480 $\times$ 800  &24  &\textbf{67.0}  &\textbf{65.5}  &\textbf{59.5} &\textbf{64.0} \\
    \rowcolor[gray]{.9} ~ & SQD-MapNet (Ours) &V2-99  &900 $\times$ 1600  &110  &\textbf{75.5}  &\textbf{74.9}  &\textbf{75.2} &\textbf{75.2} \\
    \bottomrule
  \end{tabular}
  \caption{Comparision with SOTAs on nuScenes~\cite{caesar2020nuscenes} at both 30 $m$ and 50 $m$ ranges. The $^\dag$ indicates that we have added tricks such as dcn and SyncBN to align with experimental setups of StreamMapNet~\cite{yuan2023streammapnet}. We reproduce the results of MapTR~\cite{liao2022maptr} and BeMapNet~\cite{qiao2023end} under the setting of 100 $\times$ 50 $m$ with their public codes. The - means that no corresponding results were reported in the original paper.
  }
  \label{nus}
\end{table*}

\begin{table*}
\setlength{\abovecaptionskip}{0.2cm}  
\setlength{\belowcaptionskip}{-0.3cm} 
  \centering
  \begin{tabular}{c|cccc|cccc}
    \toprule
     Range& Method & Backbone & Image Size & Epoch & AP$_{ped}$ & AP$_{div}$ & AP$_{bound}$ & mAP \\
    \midrule
    \multirow{5}*{60 $\times$ 30 \emph{$m$}} & HDMapNet~\cite{li2022hdmapnet} &Effi-B0  &-  &-  &13.1  &5.7  &37.6 &18.8 \\
    ~ & VectorMapNet~\cite{liu2023vectormapnet} &R50 &-   &-  &38.3  &36.1  &39.2 &37.9 \\
    ~ & PivotNet~\cite{ding2023pivotnet} &R50  &-  &6  &31.3  &47.5  &43.4 &40.7 \\
    ~ & StreamMapNet~\cite{yuan2023streammapnet} &R50  &608 $\times$ 608  &30  &62.0  &59.5  &63.0 &61.5 \\
    ~ & SQD-MapNet (Ours) &R50  &608 $\times$ 608  &30  &\textbf{64.9}  &\textbf{60.2}  &\textbf{64.9} &\textbf{63.3} \\
    \midrule
    \multirow{4}*{100 $\times$ 50 \emph{$m$}} & VectorMapNet~\cite{liu2023vectormapnet} &R50  &384 $\times$ 384  &120  &-  &-  &- &30.2\\
    ~ & MapTR~\cite{liao2022maptr} &R50  &608 $\times$ 608  &30  &-  &- &- &47.5 \\
    ~ & StreamMapNet~\cite{yuan2023streammapnet} &R50  &608 $\times$ 608  &30  &-  &-  &- &57.7 \\
    ~ & SQD-MapNet (Ours) &R50  &608 $\times$ 608  &30  &\textbf{66.9}  &\textbf{54.9}  &\textbf{56.1} &\textbf{59.3} \\
    \bottomrule
  \end{tabular}
  \caption{Performance comparison of various methods on Argoverse 2 \cite{wilson2023argoverse} at both 30 $m$ and 50 $m$ ranges. - means that no corresponding results were reported in the original paper. The results of VectorMapNet~\cite{liu2023vectormapnet} and MapTR~\cite{liao2022maptr} in 100 $\times$ 50 $m$ are directly borrowed from StreamMapNet~\cite{yuan2023streammapnet}. 
  Since PivotNet~\cite{ding2023pivotnet} trains the model on the full training set, we reimplement it by training with the same number of iterations for a fair comparison. 
  }
  \label{av2}
\end{table*}

\subsection{Experimental Settings}

\noindent\textbf{Datasets.} We evaluate SQD-MapNet on two competitive and large-scale datasets, \ie, nuScenes~\cite{caesar2020nuscenes} and Argoverse2~\cite{wilson2023argoverse}. The nuScenes dataset is annotated with 2Hz and each sample comprises 6 synchronized cameras. The Argoverse2 is annotated with 10Hz. Each frame contains 7 ring cameras and 2 stereo cameras. We adopt images from the ring cameras only and unify the frame rate of the dataset to 2Hz following the implementation in~\cite{yuan2023streammapnet, ding2023pivotnet}.

\noindent\textbf{Evaluation Metrics.} For the sake of fair comparison, we focus on 3 static map categories, namely \textit{lane-divider, ped-crossing}, and \textit{road-boundary}. We evaluate the models on both small perceptual range (30$m$ front and back, 15$m$ left and right) and larger perceptual range (50$m$ front and back, 25$m$ left and right). The distinct thresholds to calculate the AP is set to \{0.5$m$, 1.0$m$, 1.5$m$\} for the 30$m$ range, and \{1.0$m$, 1.5$m$, 2.0$m$\} for the 50$m$ range.

\noindent\textbf{Implementation Details.} We adopt ResNet-50~\cite{he2016deep} as backbones and use BEVFormer~\cite{li2022bevformer} with a single encoder layer for BEV feature extraction. The sizes of BEV feature map are 100 $\times$ 50 $m$ for the small perceptual range and 200 $\times$ 100 $m$ for the larger range. The strategy of streaming training is consistent with StreamMapNet~\cite{yuan2023streammapnet}. During the single-frame training phase, we adopt the normal query denoising instead of stream query denoising. All models are trained for 24 epochs on the nuScenes dataset and 30 epochs on Argoverse 2 dataset. 
We adopt AdamW optimizer~\cite{loshchilov2018fixing} with a learning rate of 5 $\times$ 10$^{-4}$.
All experiments are conducted on 8 NVIDIA Telsa V100 GPUs with a batch size of 32.

\subsection{Comparisons with State-of-the-arts}

\noindent\textbf{Performance on nuScenes.} We first compare the proposed SQD-MapNet with previous competitive vision-based counterparts on the nuScenes validation set for both 30$m$ and 50$m$ perception ranges. 
As shown in \cref{nus}, SQD-MapNet outperforms existing approaches under different perceptual range settings by a significant margin. Specifically, SQD-MapNet achieves 63.9 and 64.0 mAP within only 24 epochs under the short- and long-range evaluation settings, surpassing the previous state-of-the-art method, StreamMapNet, by more than 3.0 mAP. 
Notably, armed with the strong V2-99~\cite{lee2019energy} backbone pretrained on DD3D~\cite{park2021pseudo}, our SQD-MapNet achieves 74.0 and 75.2 mAP, setting new state-of-the-art for the competitive nuScenes benchmark. We also provide the results on the new split of the dataset adopted in StreamMapNet~\cite{yuan2023streammapnet} in the \emph{supplementary material}.



\noindent\textbf{Performance on Argoverse2.} In addition to the widely-adopted nuScenes dataset, we also gauge SQD-MapNet on the large-scale dataset Argoverse2~\cite{wilson2023argoverse} to further validate the effectiveness of our approach. To keep consistent with the evaluation protocol in nuScenes dataset, we report the results for both 60 $\times$30 $m$ and 100 $\times$ 50 $m$ ranges. As shown in \cref{av2}, SQD-MapNet consistently outperforms StreamMapNet by about 2.0 mAP, validating the generalization and superiority of our approach. For the new split of dataset adopted by StreamMapNet~\cite{yuan2023streammapnet}, we additionally show the results in the \emph{supplementary material}.


\subsection{Ablation Study}
In this section, we provide extensive ablation studies to explore the effectiveness of main components in SQD-MapNet, providing a deeper understanding of our approach. If not specified, all experiments are conducted on nuScenes dataset at a perceptual range of 60 $\times$ 30 $m$.

\noindent \textbf{Main Ablations}
To understand how each component contributes to the final performance, we subsequently add the proposed modules to our baseline and report the performance in \cref{tab:main_ablations}. 
We regard StreamMapNet~\cite{yuan2023streammapnet} without temporal streaming as the vanilla baseline. When applying the streaming strategy proposed by~\cite{yuan2023streammapnet}, the performance declines by 0.7 mAP, indicating that it is difficult for the network to learn the constant changes in curves of different frames. Then we add the dynamic query noising mechanism in the training process, the mAP is improved by a large margin (3.7 mAP). Additionally, when adaptive temporal matching is applied, we observe a further 1.0  improvement, which verifies the necessity of the matching strategy. Overall, the performance of SQD-MapNet achieves 63.9  mAP, yielding a performance enhancement of 4.0  mAP.

\begin{table}[!h]
\setlength{\abovecaptionskip}{0.2cm}  
\setlength{\belowcaptionskip}{-0.6cm} 
    \centering
    \begin{tabular}{l|c}
        \toprule
         Method & mAP\\
         \midrule
          Single-frame baseline &59.9 \\
          + Temporal Stream &59.2 (-0.7) \\
          + Dynamic Query Noising &62.9 (+3.7) \\
          + Adaptive Temporal Matching &63.9 (+1.0) \\
        \bottomrule
    \end{tabular}
    \caption{Ablations  on  each component in SQD-MapNet. Starting from a single-frame baseline to the final model.}
    \label{tab:main_ablations}
\end{table}



\noindent \textbf{Ways for Adaptive Temporal Matching}
~\cref{tolerance} ablates the performance of 
different matching ways and mathcing scales on SQD-MapNet. Concretely, $\alpha$ denotes the value of adaptive matching scale and $\beta$ is the predefined fixed matching threshold.
Due to the ignorance of the curve's properties, a fixed threshold only yields a maximum score of 62.8 mAP. When $\alpha$ is small, the result is not optimal, suggesting that strict matching may cause some ground truths to be filtered out during the denoising process. As the value of $\alpha$ increases, which means the tolerance level of transformation bias becomes larger, more positive samples of the previous frame can be matched. Specifically, the performance reaches 63.5 mAP when $\alpha$ equals 0.1. However, when $\alpha$ keeps increasing, the detection accuracy starts to incline, indicating that there are plenty of noisy samples from the previous frames incorrectly matched with ground truths at the current frame.

\begin{table}[!h]
\setlength{\abovecaptionskip}{0.2cm}  
\setlength{\belowcaptionskip}{-0.6cm} 
    \centering
    \begin{tabular}{cc|c}
        \toprule
         Matching Scale & Adaptive Matching & mAP\\
         \midrule
          $\beta$ = 2.5 & \xmark & 62.0 \\
          $\beta$ = 4.0 & \xmark & 62.3 \\
          $\beta$ = 5.0 & \xmark & 62.8 \\
         $\alpha$ = 0.05 & \cmark & 63.0 \\
          $\alpha$ = 0.1 &  \cmark & \textbf{63.5} \\
          $\alpha$ = 0.2 & \cmark & 62.9 \\
          $\alpha$ = 0.3 & \cmark & 63.0 \\
        \bottomrule
    \end{tabular}
    \caption{Comparison of matching ways and matching scale. }
    \label{tolerance}
\end{table}

\noindent \textbf{Decay Rate of Noise}
After warping the previous frame curve to the current frame, there is a natural noise between the previous one and the current ground truth. Besides, we add extra random noise during the dynamic query noising phase. Therefore, deciding the right level of decay rates of the added noise is non-trivial. We experiment SQD-MapNet with different $\gamma$ in \cref{decay_rate}. From the table, we can conclude that a small ratio of 0.2 can balance both the learning diversity and the negative effect introduced by noise. 

\begin{table}[!h]
\setlength{\abovecaptionskip}{0.2cm}  
\setlength{\belowcaptionskip}{-0.6cm} 
    \centering
    \begin{tabular}{c|cccc}
        \toprule
         Decay Rate &AP$_{ped}$ & AP$_{div}$ & AP$_{bound}$& mAP\\
         \midrule
          $\gamma$ = 0.1 &62.5&65.3& 62.5&63.4 \\
          $\gamma$ = 0.2 &\textbf{63.0}&65.5&63.3 &\textbf{63.9} \\
          $\gamma$ = 0.3 &59.0&64.1&\textbf{63.8} &62.3 \\
          $\gamma$ = 0.5 &59.7&\textbf{65.8}&63.7 &63.0 \\
          $\gamma$ = 0.7 &61.7&64.0& 61.3&62.4 \\
        \bottomrule
    \end{tabular}
    \caption{Effect of different decay rates of noise on the final results.}
    \label{decay_rate}
\end{table}

\begin{figure*}[h]
\setlength{\abovecaptionskip}{0.2cm}  
\setlength{\belowcaptionskip}{-0.2cm} 
  \centering
   \includegraphics[width=1.0\linewidth]{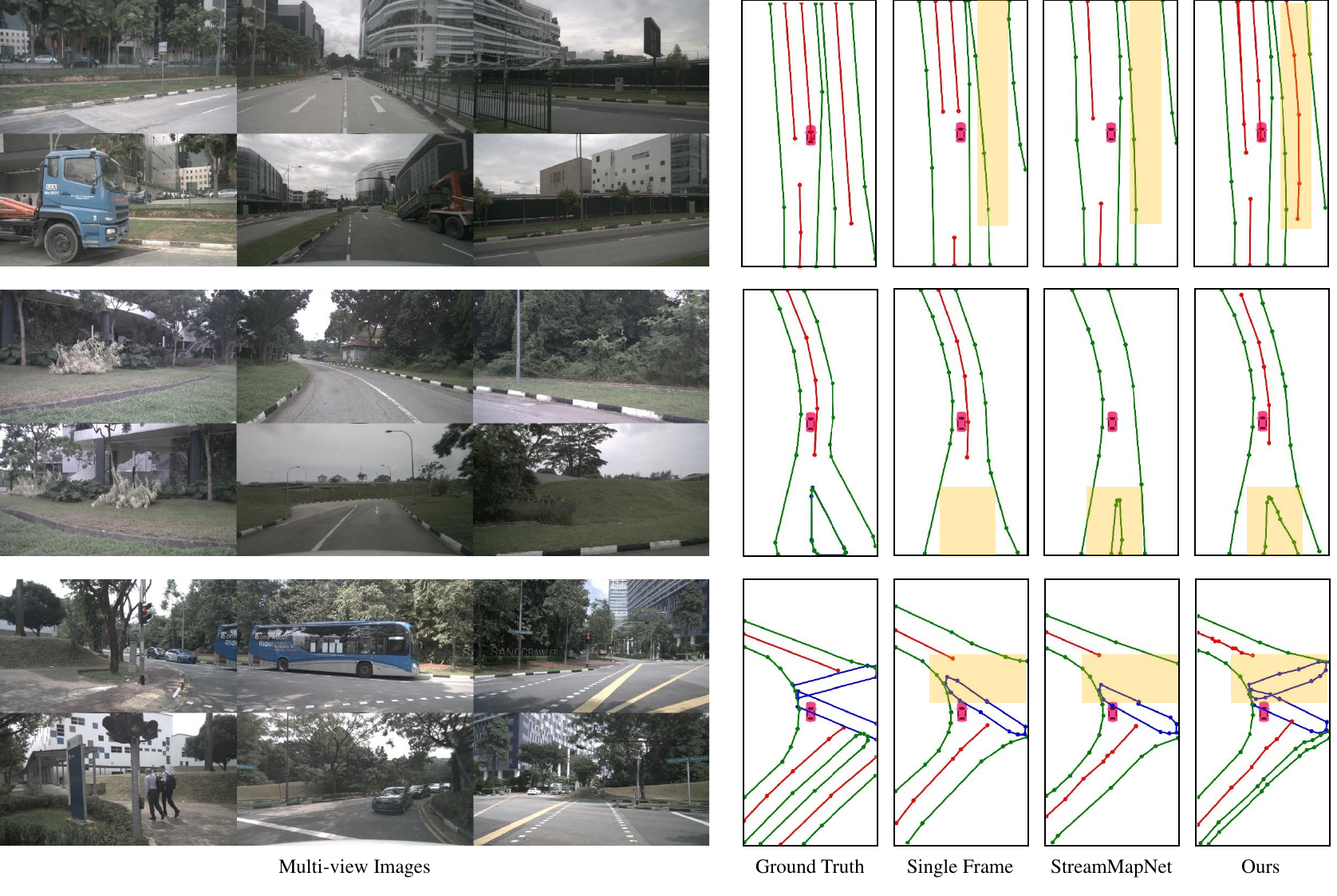}
   \vspace{-15pt}
   \caption{Comparison with the single-frame model and StreamMapNet~\cite{yuan2023streammapnet} on qualitative visualization under different scenarios.  In the HD-map, green lines denote \textit{road boundaries}, red lines indicate \textit{lane-dividers}, and blue lines denote \textit{pedestrian crossings}.
}
   \label{fig:vis}
\end{figure*}

\noindent \textbf{Normal and Stream Query Denoising}
We also compare the performance of our baseline, normal query denoising, and stream query denoising in ~\cref{dn_streamdn}. It shows that normal query denoising improves 3.5 mAP performance compared to the baseline. Stream query denoising further outperforms the normal query denoising by 1.2 mAP. The reasons can be two-fold: (i) A portion of temporal curves are identical to the ground truth of the current frame, so normal query denoising is sometimes equivalent to stream query denoising, which enhances the ability to model stream queries. (ii) stream query denoising helps the transformer decoder learn to exploit the stream queries by introducing temporal ground truth.

\begin{table}[!h]
\setlength{\abovecaptionskip}{0.2cm}  
\setlength{\belowcaptionskip}{-0.6cm} 
    \centering
    \begin{tabular}{c|c}
        \toprule
         Way of Denoising & mAP\\
         \midrule
          w/o Denoising &59.2 \\
          Normal Query Denoising &62.7 \\
          Stream Query Denoising &\textbf{63.9} \\
        \bottomrule
    \end{tabular}
    \caption{Comparison of denoising and temporal denoising.}
    \label{dn_streamdn}
\end{table}

\subsection{Qualitative Analysis}
We show some qualitative comparisons with a single-frame model, StreamMapNet~\cite{yuan2023streammapnet}, and SQD-MapNet in \cref{fig:vis}. The single-frame model is reproduced with ~\cite{yuan2023streammapnet} without temporal information. We can find that the single-frame model and StreamMapNet easily fails to recognize some curves in both simple and complex scenes. 
Compared with other methods, SQD-MapNet utilizes the temporal information more effectively and accurately recognizes curves. 

\section{Conclusion}
\label{sec:conclusion}
In this paper, we introduce the stream query denoising (SQD) strategy to enhance the temporal streaming modeling in HD-map construction. 
Such strategy is only applied during training process and can be removed for the inference. Extensive experiments on nuScenes and Argoverse2 show that the performance of streaming models (e.g., StreamMapNet) are greatly improved with the proposed SQD strategy. The resulting SQD-MapNet framework is remarkably superior to other existing methods across all settings of close range and long range. 

\noindent \textbf{Discussion:} 
Our proposed SQD strategy greatly enhances the temporal modeling of HD-map construction, as verified on nuScenes and Argoverse2. However, the effectiveness of our method can only be verified based on  streaming models. And there is only one streaming baseline StreamMapNet available at present, lacking of abundant verification. 
The temporal modeling on HD-map construction is quite important for autonomous driving. It can keep the temporal consistency and stability of predictions cross different frames. For example, temporal modeling can deal with the case where the confidence score of prediction is floated around the set threshold.
Therefore, related researchers in the community are encouraged to explore more on the temporal modeling for HD-map construction. The temporal consistency, achieved by our adaptive temporal matching may provide some new insight for the community. 


{
    \small
    \bibliographystyle{ieeenat_fullname}
    \bibliography{main}
}


\end{document}


\maketitle  

\appendix

\section*{Appendix}

\section{More Detailed Implementation Settings}
The weights of the classification loss, the line loss and the translation loss are 4.0, 50.0, and 0.1, respectively. The weights of the denoising classification loss and the denoising line loss are set to 4.0 and 50.0. The number of denoising queries is 60. Moreover, for the noise setting of denoising queries, we change the original category labels with a probability of 50\%, and we set it to 0.6 for the maximum noise scale of the position.

\section{More Experiments On New Split}
\label{sec:exper}
StreamMapNet~\cite{yuan2023streammapnet} found an overlap of over 54\% locations between the training and validation sets, including nuScenes~\cite{caesar2020nuscenes} and Argoverse2~\cite{wilson2023argoverse}, and therefore proposed a new way to split the datasets. We provide results for the two datasets at the 30 $m$ perceptual range under the new split.

\noindent\textbf{Performance on nuScenes.} We first compare the proposed SQD-MapNet with previous competitive vision-based counterparts on the nuScenes new validation set. 
As shown in \cref{nus}, SQD-MapNet outperforms existing approaches on new split and achieves 35.9 mAP within only 24 epochs. 

\begin{table}[h]
  \centering
    \scalebox{0.95}{
  \begin{tabular}{c|cccc}
    \toprule
     Method & AP$_{ped}$ & AP$_{div}$ & AP$_{bound}$ & mAP \\
    \midrule
    VectorMapNet~\cite{liu2023vectormapnet} &15.8  &17.0  &21.2 &18.0 \\
    MapTR~\cite{liao2022maptr}  &6.4  &20.7  &35.5 &20.9 \\
    StreamMapNet~\cite{yuan2023streammapnet}  &29.6  &\textbf{30.1}  &41.9 &33.9 \\
    SQD-MapNet (Ours)  &\textbf{33.7}  &29.5  &\textbf{44.5} &\textbf{35.9} \\
    \bottomrule
  \end{tabular}
  }
  \caption{Comparision with SOTAs on the new nuScenes~\cite{caesar2020nuscenes} split at 30 $m$ perception range. The results of VectorMapNet~\cite{liu2023vectormapnet}, MapTR~\cite{liao2022maptr} and StreamMapNet~\cite{yuan2023streammapnet} are directly from~\cite{yuan2023streammapnet}. 
  }
  \label{nus}
\end{table}

\noindent\textbf{Performance on Argoverse2.}
As shown in \cref{av2}, SQD-MapNet outperforms existing approaches by a significant margin. Specifically, SQD-MapNet achieves 61.2 mAP within only 30 epochs, surpassing the previous state-of-the-art method, StreamMapNet~\cite{yuan2023streammapnet}, by more than 3.0 mAP. 

\begin{table}
  \centering
  \scalebox{0.95}{
  \begin{tabular}{c|cccc}
    \toprule
     Method & AP$_{ped}$ & AP$_{div}$ & AP$_{bound}$ & mAP \\
    \midrule
     VectorMapNet~\cite{liu2023vectormapnet}  &35.6  &34.9  &37.8 &63.1 \\
    MapTR~\cite{liao2022maptr}  &48.1  &50.4  &55.0 &51.1 \\
    StreamMapNet~\cite{yuan2023streammapnet} &56.9  &55.9  &61.4 &58.1 \\
    SQD-MapNet (Ours)  &\textbf{61.2}  &\textbf{58.1}  &\textbf{64.3} &\textbf{61.2} \\
    \bottomrule
  \end{tabular}
  }
  \caption{Comparision with SOTAs on the new Argoverse2~\cite{wilson2023argoverse} split at 30 $m$ perception range. The results of VectorMapNet~\cite{liu2023vectormapnet}, MapTR~\cite{liao2022maptr} and StreamMapNet~\cite{yuan2023streammapnet} are directly from~\cite{yuan2023streammapnet}. 
  }
  \label{av2}
\end{table}

{
    \small
    \bibliographystyle{ieeenat_fullname}
    \bibliography{main}
}
